# Media Forensics and Deepfake - Systematic Survey

**Nadeem Jabbar CH**[*1] **Aqib Saghir**[2] **Ayaz Ahmad Meer**[2]
**Salman Ahmad Sahi**[2] **Bilal Hassan**[3] **Siddiqui Muhammad Yasir**[4]

[1]Department of Computer Science, The Superior University, Lahore, Pakistan
nadeem.ch@superior.edu.pk,
[2]Department of Computer Science, The Superior University, Lahore, Pakistan,
[3]Northumbria University, London Campus, UK,
[4]School of Artificial Intelligence, Tongmyong University, Busan, South Korea

ResearcherStore

**ABSTRACT**

Deepfake is a generative deep learning algorithm that creates or changes facial features in a very realistic way, making it hard to differentiate the real from the fake features. It can be used to make movies look better, as well as to spread false information by imitating famous people. In this paper, many different ways to make a Deepfake are explained, analyzed and separated categorically. Using Deepfake datasets, models are trained and tested for reliability through experiments. Deepfakes are a type of facial manipulation that allow people to change their entire faces, identities, attributes, and expressions. The trends in the available Deepfake datasets are also discussed, with a focus on how they've changed. Using Deep learning, a general Deepfake detection model is made. Moreover, the problems in making and detecting Deepfakes are also mentioned. As a result of this survey, it is expected that the development of new Deepfake-based imaging tools will speed up in the future. This survey gives in-depth review of methods for manipulating images of face, and various techniques to spot altered face images. Four types of facial manipulation are specifically discussed which are attribute manipulation, expression swap, entire face synthesis, and identity swap. Across every manipulation category, we yield information on manipulation techniques, significant benchmarks for technical evaluation of counterfeit detection techniques, available public databases, and a summary of the outcomes of all such analyses. From all of the topics in the survey, we focus on the most recent development of Deepfake, showing its advances and obstacles in detecting fake images.

# 1 Introduction

Many deep learning-based methodologies are used to generate temporal data and fake spatial by changing one person's face (posture, identification, expression) and so on with another is referred to as the famous term "Deepfake." Digitally altered photos and videos, especially with Deepfake technologies, have been a major source of public concern. This technology was being used for fake news, fake pornography, hoaxes, and financial fraud, etc.[1]

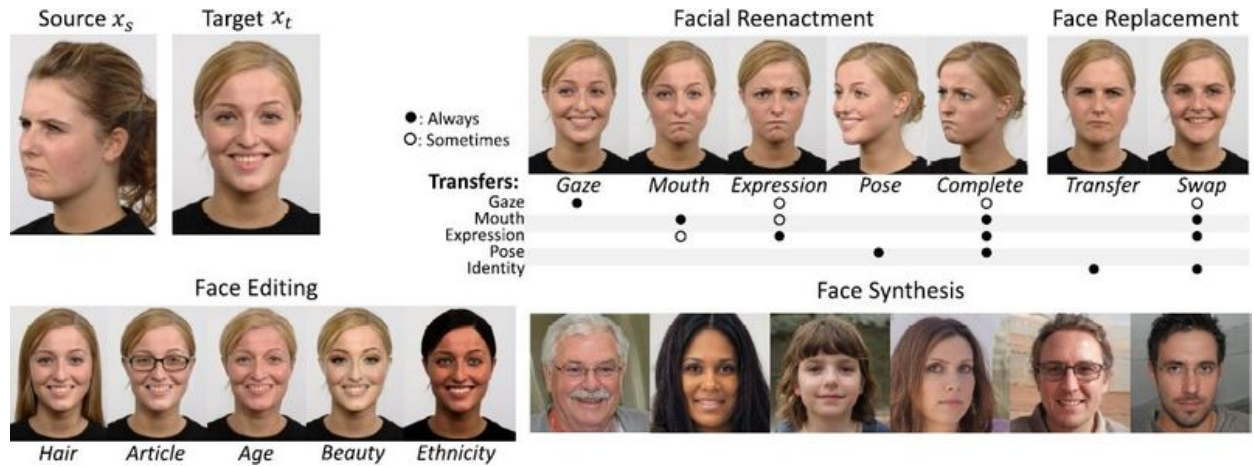

**Fig. 1.** Different categories of deepfake Data [1]

However, many things changed dramatically in the recent years, starting with the early works and continuing to the present. The ability to mechanically synthesize non-existent faces or modify the genuine face of a single person in an image or video is becoming increasingly common these days. Before Deepfake, fake media generation required expertise in the field, and took more time and effort. Tools like Adobe Lightroom and Adobe Pohtoshop are used by experts for retouching and modifying a picture[2]. Deepfakes, which are now generated using artificial intelligence, have no need for too much expertise. Fake data can be generated only in a few clicks. In graphics industries and computer vision Deepfake has a wide range of benefits as well as harmful uses; with the help of Deepfake stunning landscapes can be generated; on the other hand, fake news generation can mislead the audience, which may increase security risks. To generate fake movies of female celebrities, many people with malicious motives have used these technologies in such a way that triggered important societal challenges, including sexual harassment. Furthermore, several malicious applications, like DeepNude4[3], have taken advantage of Deepfake it allows them to take a snapshot of a fully suited lady and produce an image in which her clothing are removed. Since a few years, researchers have been working on developing various techniques to distinguish images and videos that are synthesized with artificial intelligence, especially those manipulated with GANs and their variants) from those that are captured naturally with a camera. There are competition such as DFGC ( Deepfake game challenge)[4] encourage researchers to propose better solution for detection. Detecting fakes in the spatial domain is one of the most commonly used strategies for finding fake videos and images in recent studies [2]. In the frequency domain, several researchers are attempting to distinguish between original and forged data, which is a relatively new field[2] Deepfake creation is classified into four types: 1. Facial Reenactment 2. Face Replacement 3. Face Editing 4. Face synthesis

In this survey, Deepfake detection strategies for images and videos are gone through. Moreover, the Deepfake creation processes and datasets that are used to detect Deepfakes, as well as the Deepfake detection algorithms, are discussed.

Below are the major objectives of this article, which are summarized as follows:

- Deepfake tools, which are used to change many features of photos and videos, are undergone.
- Deepfake datasets, as well as certain standard datasets, are introduced for forensic review.
- Some of the most recent methods for detecting Deepfake in images and videos are mentioned.
- The rest of the article is decomposed as: Paper collection and review methods are provided in 2 . Then 3 is about deepfake creation. 4 is outlined with Public Datasets. However, in 5, Deepfake detection is focused. Finally, discussions and future work are drawn in 6.

## 2 Prepare Collection And Review Methods

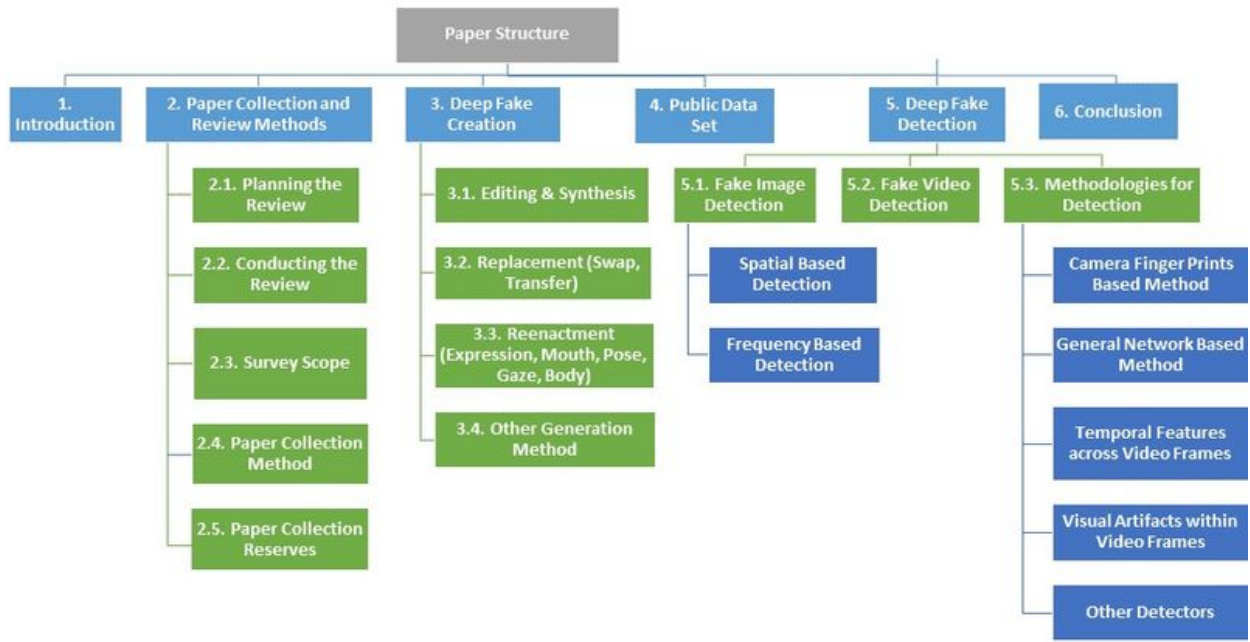

**Fig. 2.** Tree diagram showing structure of the paper

## 2.1 Planning the Review

The primary objective of this step is to deduce the problem that needs to be solved. Next, to come up with the idea through which the problem can be solved, and evaluate or solve the problem. In this step, the idea and its requirements are provided. Eventually, the final step is to test the reliability of the criteria and procedures used to solve the problem.

## 2.2 Conducting the Review

It includes six obligatory steps on the basis of the guiding concepts outlined in this phase.

**Research Questions (RQs)**

Because the proper question leads to the right direction and gives the study significance. Therefore, the success of the study is dependent on the appropriate and relevant research questions[5]. So, we outline the following research questions: 1) The first question is, what are the most famous Deepfake detection methods? 2) What datasets are typically utilized for detecting Deepfake? 3) What are the features typically utilized in detecting Deepfake? 4) What models are used to detect Deefake manipulation? 5) For Deepfake Detection Techniques, what is the Classification Structure? 6) What is the overall productivity of various Deepfake detection techniques on the basis of experimental evidence?

### Search Strategy

In this stage, the maximum possible relevant data is gathered. In order to detect Deepfake, this article includes as many combinations of related search phrases as possible. Using Boolean terminology, search terms were combined with "AND" or "OR," which is the main idea. The search terms are: Deepfake AND Digital Media Forensics, FaceSwap OR Deepfake, Video manipulation OR Fake image/video/face detection OR Facial Manipulation. For research purposes, more than one or two sources were used. Furthermore, the authenticity of the research was tested thoroughly using multiple resources as mentioned below. The numerous research articles are available online, the top repositories that were both important and easy to use were selected which are shown

- ScienceDirect (ELSEVIER)
- ACM Digital Library
- Web of Science
- Google Scholar
- Database Systems and Logic Programming (DBLP)
- SpringerLink
- Computing Research Repository
- IEEE Xplore Digital Library
- Semantic Scholar

### Selection Criteria

To find the most reliable articles, three rules were set up for this research.

- The keywords, title, or abstract must contain the search phrases (Some articles, although based on Deepfake, did not contain words like "Deepfake" in the keywords, title, or abstract.)
- In writing, empirical evidence is made clear.
- The main goal of this study is about how images or videos can be changed and how we can detect change images and videos

### Quality Assessment Criteria (QAC)

Analyzing the material in the document and evaluating the quality of the evidence are equally important. Researchers should be cautious when interpreting the findings of a poorly executed study since they may have been influenced by methodological flaws. For that reasons, right criteria must be chosen in order to assess the quality of the evidence. We validate the selected studies using the criteria defined in[6] and evaluate the gathered material by applying these criteria.

In our quality assessment phase, we decided 96 research articles and 21 additional reviews (7 SLRs, 10 analyses, and 4 surveys) relating to Deepfake detection.

**Data Synthesis** The goal of data synthesis is to arrange and compile the findings of the conducted research. We evaluate the data, once accumulated, in order to extract additional information. Also, we use various tools for data visualization, including tables and charts, to display the acquired data.

**Reporting the Review** After reviewing all of the research, we organized the results properly for the target readers and distribution medium.

## 2.3 Survey Scope

The technical aspects of Deepfakes are the paper's primary focus. This paper briefly discussed spotting a deepfake and avoiding being caught, as different research publications cover it. The moral and social elements of Deepfakes, which are discussed in this survey, are not the primary focus of this study. There are several types of media manipulation, including body, expression, and tone. However, we focused only on Deepfakes of facial in our survey.

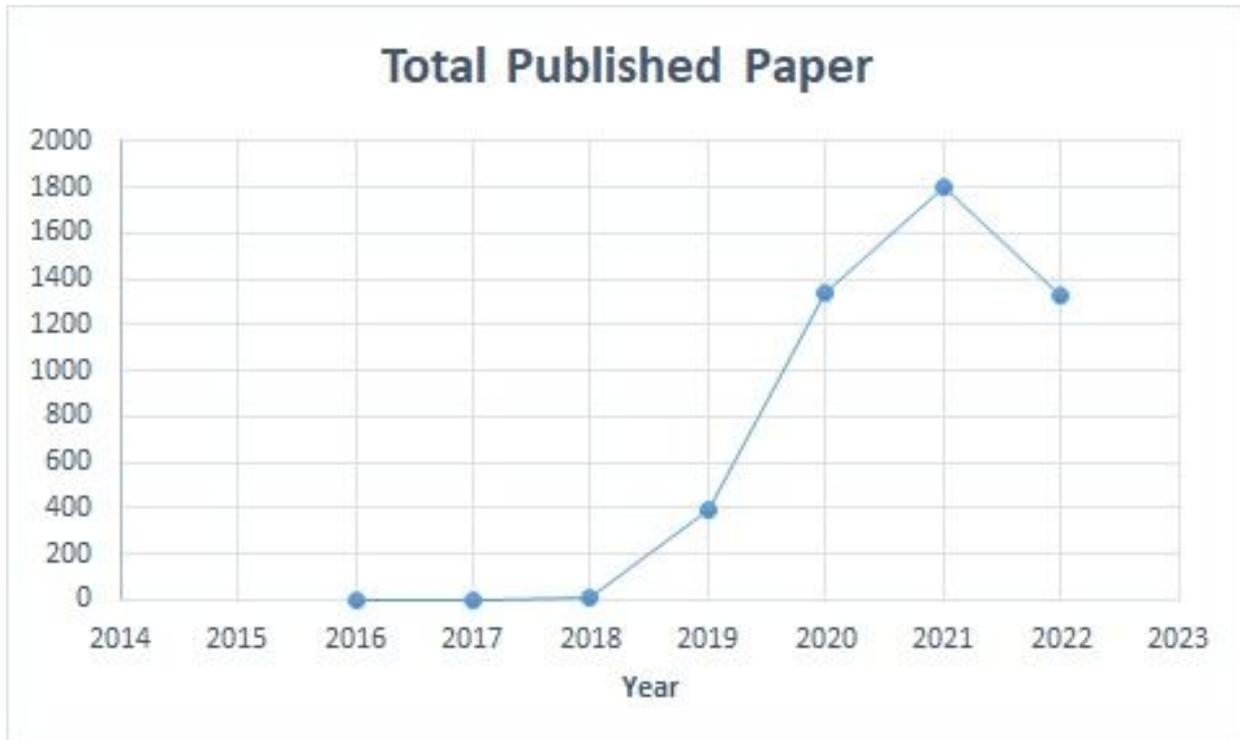

**Fig. 3.** Number of published paper from year 2015 to 2022 as per Google Scholar and **[7]**

## 2.4 Paper Collection Method

First, over 100 papers on Deepfakes generation, Deepfakes detection, and Deepfakes evasion were gathered from arxiv. Next, on two major scientific databases (IEEE, DBLP, Google Scholar, and various other sources from which the latest research publish) we used keywords in searching to recognize and collect papers on Deepfake. To ensure a more comprehensive and accurate survey, publications in major conferences and respective workshops were also browsed of last three years to ensure a more thorough and precise survey. The following keywords are used: video synthesis, tampered face, facial image, face swapping, deepfake, GAN-synthesized, , forgery, and AI-synthesized - manipulation.

## 2.5 Paper Collection Reserves

We found 287 papers from arXiv, Google Scholar and IEEE. The papers were mostly about how to make Deepfake, Deepfake detection, and how to avoid detection of Deepfakes. Figure 3illustrates the allocation of research that was released in various locations. Here, papers from the top journals and sources were divided into several groups. The "Others" category included publications from lesser-known publications and conferences. Furthermore, Unpublished papers were categorized as "arXiv".

# 3 Deepfake Creation

In recent years, Computer vision has progressed and achieved excellent results through deep learning. Deep learning is also one of the most widely used technologies for creating artificial content. Meanwhile, Digital images and videos manipulation is one of the leading interests in deep learning. A wide range of workshops and conferences have a special session for Deepfake, which reflects the interest of the research community [7]. Furthermore, different challenges like Media forensics Challenge (MFC 2018) and Deep Fake Detection Challenge (DFDC) have been launched, which also reflects the significance of the topic.

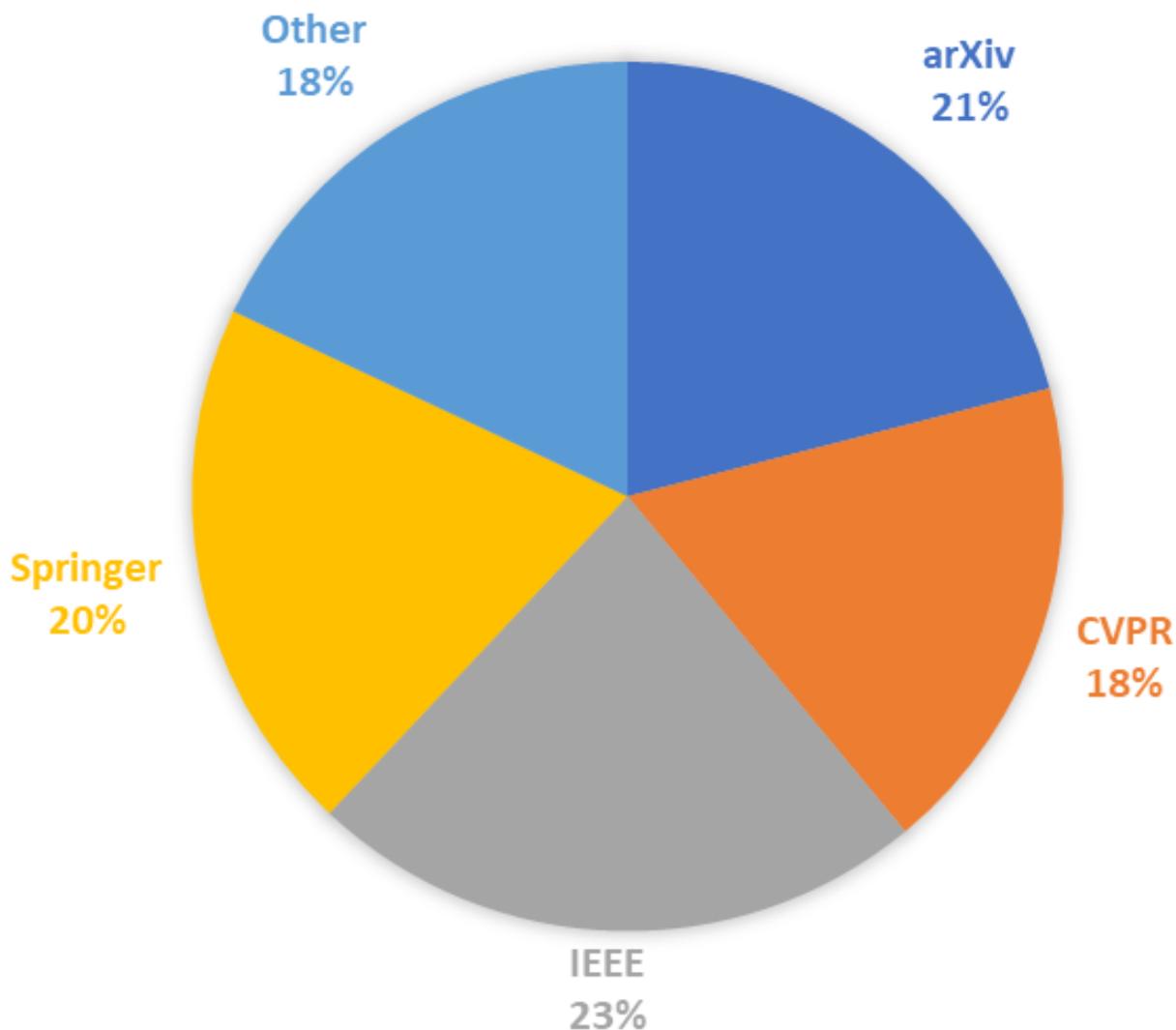

**Fig. 4.** Number of papers collected for the current study

The technology to create Deepfakes has existed for decades; however, creating these effects took an entire studio of experts with just a short time ago. Currently, Deepfake technologies are capable of manipulating pictures or videos much more swiftly due to the addition of Artificial Intelligence (Deepfake means AI, Deep learning base content). GAN's (Generative Adversarial Networks) is the most known source that provides unbelievably realistic results in deep learning[8]. An undeniable fact is that GANs are tough to work with and need a lot of training data. [9]There are many challenges in the development of GAN while creating smooth and manageable syntheses, mostly focused on the high-resolution domains. On top of that, there are many mobile and desktop-based applications in the market that assist people in creating Deepfakes. Each application uses its own developed GAN to achieve its purposes. ZAO, a mobile application[10], AutoFaceswap[11] and FaceApp[12] help the general public on the internet to create Deepfakes without any difficulties. Face swapping is a highly used technique in the field of Deepfake. Through this technique, one face can be moved from the source image to the target image and the main idea behind this technique is also GANs. Over time, images created by Style-GAN[13] and its variants such as Style-Gan2[14] and Style-GAN-Ada[15]are getting ever more lifelike and utterly unrecognizable to the naked human eye. Each manipulation method has its unique effect, such as without modifying any other facial traits changing eye size or altering skin color. In contrast, it is impossible to generate HD human faces using Style-GAN[13]. On the other hand, BigGAN[16] cannot manipulate the length and complexion of a person without modifying other features of images.

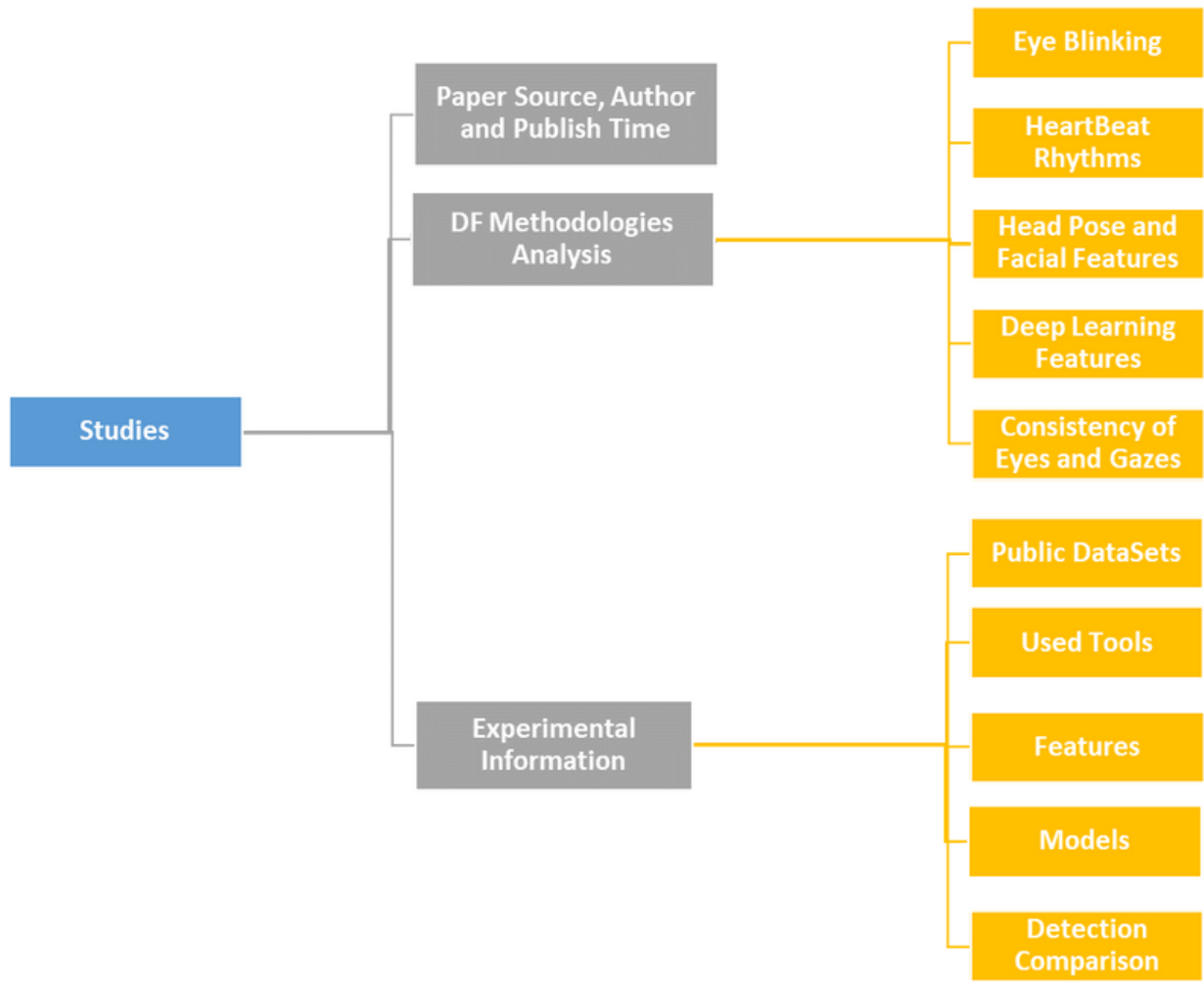

**Fig. 5.** Flow chart of explained information in current study.

Below, we cover the categories of manipulation in more detail.

## 3.1 Editing

In the enchantment, Deepfake, the features of the target image can be changed, removed or added. The technology can be used to manipulate facial hair, weight, age and clothes[17] etc. FaceApp[12]is widely used for editing that is done just for entertainment purposes. However, it can also be used falsely to attack a target personality. Due to this, a constantly growing concern has settled related to the developing of Deepfake mechanics.This technology has made it possible to produce proofs of the actions that never happened. For instance, (i)an ill leader could be made to appear well.[18]. (ii) remove object's clothes for the entertainment and blackmailing purpose[19] etc.

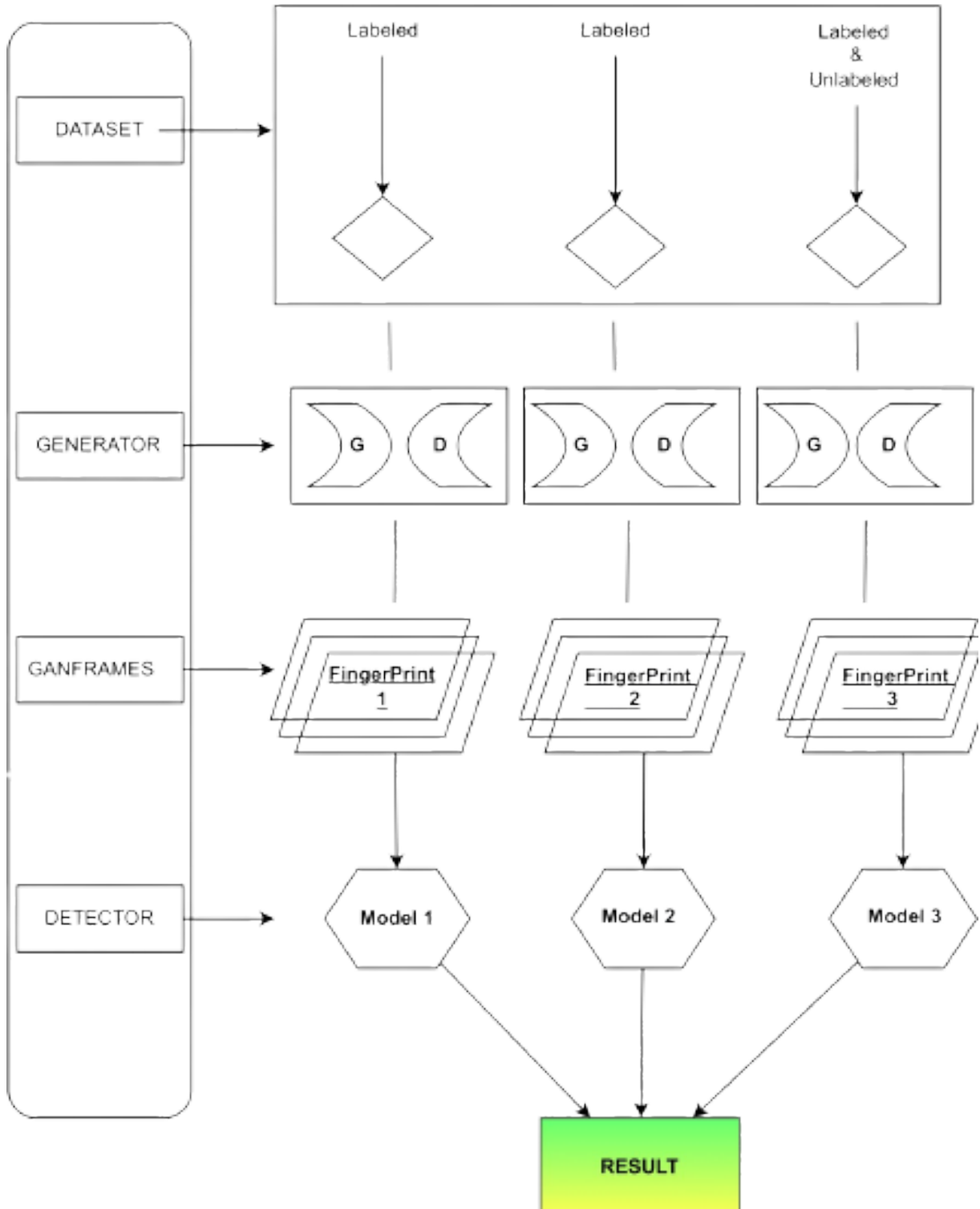

**Fig. 6.** GAN fingerprint example, that can be found in Deepfake-generated media

## 3.2 Synthesis

Synthesis method uses extremely intelligent GANs such as Style-GAN[13] to generate faces which do not exist in the form of images. This method generates amazing outputs such as facial images of high-level resolution. It can be used in different fields such as video games, movies. Yet, like the editing method, synthesis can also be used for negative purposes.

## 3.3 Identity Replacement

It is another trait of deep learning. When the content of the target is replaced by the source while taking care of identity, it is called identity replacement. The following are the two methods used for identity replacement.

**Swap**

Identity swap is quite common in the world of Deepfake. This method is also commonly known as face swap. It can be generated by transferring content to target from source which is driven by target. In simple words, one can replace only a few features of the face like open eyes OR smile persons digitally.

**Transfer**

In the transfer technique output is generated when the content of the source image is exchanged with that of the target image. The fashion business frequently uses facial transfers to visualize people wearing various outfits.

**Identity Replacement or attack model**

It is well known and widely used in negative activities and applications[20]. For example, an attacker can swap the face of the victim with a naked porn-stars in order to defame and blackmail the victim. This strategy has previously been employed as a means of disseminating political viewpoints.[21]

## 3.4 Reenactment

Another manipulation technique for creating Deepfakes is reenactment. Based on GAN architectures, the available techniques like neural textures[22] and Face2Face[23]are used to make a video, in which the facial expression of one person can be replaced by the facial expression of some other person. The following list of reenactment's branches is provided:

**Expression Swap** is best described as a face modifier technique, which is generated when source image is used to derive the expression of target. It is also the most common technique used in movies, video games, digital media etc. where the performance of the person can be enhanced and improved. Researchers altered 3D-scanned head models in 2003[24]. Without a 3D model, it was proved that it was possible in 2005. Eventually, between 2015 and 2018, Justus Thies et al. [23][25][26] demonstrated the use of 3D parametric models to provide high-quality, in-the-moment outcomes. In 2022, S.M. Yasir [27]proposed a method to detect 3D instance from 3D point cloud that is valuable for 3D face detection and its applications.

**Mouth reenactment** has been used for many years. This method is recognized as dubbing in which the mouth of the target person is derived by the source or an audio input containing conversation. For long, this technology is being widely used when dubbing a movie into another language. For example, if a movie is originally directed in Spanish Language, it can be dubbed into English language using this method. The developers of Obama Net[28] proposed a network that uses text as input rather than audio to reproduce a person's mouth and voice.

**Gaze reenactment** is another kind of manipulation approach generated when the direction of target eyes and eyelids are driven by the source. This procedure is generally used to get realistic output of photographs to maintain eyes contact[29] or to change the direction of one's gaze. The Gaze Redirection Network was proposed by Y. Yu et al. [30] (GRN). In GRN, the source angle, head pose, and cropped eye of the target are individually encoded before being sent via an encoder-decoder network to produce an optical flow field. The generator is driven by an optical flow field that is applied to successive frames in a source video.

**Pose reenactment** is a commonly used technique of manipulation generated when the head position of the target is derived by the source. This method has mainly been used for scanning the faces of individuals in a security footage. Moreover, this approach can be used as a means for improving facial recognition software[31]. J. Cao et al.[32], The authors recommend employing two GANs: The first frontalis the face to produce a Ultraviolet map, [27] and the second turn around the face at the required target angle. As a result, each model executes a simpler procedure. Thus, the models work together to create an image of superior quality.

**Body reenactment**, which is also known as pose transfer and human pose synthesis which is a branch of reenactment approach that are listed above and except that it is pose of source body transfer to the target body.

Table 1: Facial manipulation techniques used to create Deepfakes

| Table 1 | |
|---|---|
| **Facial Manipulation** | **Key Idea** |
| Entire Face Synthesis | Generates perfect, nonexistent facial images using a powerful GAN model, such as StyleGAN or StyleGAN2-Ada. |
| Identity Swap | Using FaceSwap or Deepfake methods to swap out a person's face for another person's face in a photograph or video. |

| Attribute Manipulation | Altering some facial features, such as the colour, age, gender, skin, hair, and eyeglasses, for example, using Star-GAN. |
| --- | --- |
| Expression Swap | Altering the facial expression of one person in a video with the facial expression of another person, e.g., Face2Face, NeuralTextures. |
| Miscellaneous | Face morphing is the process of producing synthetic biometric face samples that mirror the provided biometric data. Remove the identity information shown on a face image or video with face de-identification. Facial expression swap, sometimes known as lipsinc deepfakes, is a technique used to synthesise video from audio or text. |

## 3.5 Miscellaneous

Other than the above given categories of creating Deepfake, there are some further types of manipulation. Hereunder, we discuss three of the various evolving techniques, including Face morphing, Face de-identification, and text to video or audio to video swaps of facial expression.

**Face Morphing** generates biometric faces, which are purely artificial and are copies of biometric features from different sources. In other words, face morphing is the utilization of picture control projects to consolidate two separate pictures into a fresh brand-new picture. U. Scherhag et al.[33] investigation of the face morphing approach in 2019 demonstrates the use of attack detectors and morphing algorithms.

**Face de-identification** The primary objective is to obscure a person's identity from a facial video or image to protect their privacy [34]. There are various ways to accomplish this. The simplest method may be to blur or pixelate the face (e.g., in Street View of Maps). More advanced techniques attempt to produce face images with distinct identities while keeping all other elements (position, emotion, illumination, etc.) the same. As a result, the idea of face de-ID is relatively broad. Face identity swapping could be one method for achieving face de-identification. Applying face de-ID to still photos was the foundation of earlier studies in this field. A multifactor framework for de-ID was provided by Gross et al. in [35], including linear, bilinear, and quadratic models. On an expression variant face database, they demonstrated that their technology could safeguard anonymity while maintaining the usefulness of the data. Recent advances in image synthesis methods depend on generative deep neural networks, especially GAN, and have prompted new face de-ID techniques that substitute synthesized faces for real ones, such as those used in [36] and [37]. A further suggestion made by the authors in [38] was to confuse random face-based gender classifiers by using semi-adversarial networks (SAN). Recently, in [39], Gafni et al. revealed 2019 a technique for face de-ID with compelling performance even in unconstrained movies. Their strategy is based on a learned face classifier and an adversarial auto-encoder. They can create a rich latent space in which identification and expression information are embedded. A fresh face de-ID technique built on a deep transfer model was also given in [40]. [41]This technique interprets the non-identity-related facial characteristics as the original faces' style. It applies a trained facial attribute transfer model to extract and map those characteristics to various faces, yielding highly encouraging results in still photos and videos.

**Text to video or Audio to video** both are the branches that also come under the tree of lip-sync Deepfakes[42]. These methods create videos by synthesizing the target's expression through text or voice. Artificial Intelligence has made people capable of generating Deepfakes that they type whatever they want their subject will say. For example, these Deepfake methods are used in making three-dimensional videos, such as cartoon videos. Using these methods, the written script or the audio of a speaker can directly be converted into a video just like converting text to video. The method of generating high-quality video of someone conversing with an exact lip-sync track is explained through an example of a video[43]. In [44], [45] discuss more key state-of-the-art techniques Video portrait editing technique and Audio-based technique. Furthermore, Fried et al.[46] came up with a text-based editing approach for changing speech content in videos. This is done to create a new video by taking clips from a video of a person speaking and combining the necessary spoken material with the person's lips to match the new words.

# 4  Public Datasets

## 4.1 Real Datasets

**CelebA:** Liu et al. [47] it built through labeled images of celebrities faces. There are ten thousand celebrities in CelebA dataset, each of which has twenty images. This means there are 200,000 images in total. A professional labelling company annotates every photo in CelebA with and five key points and forty face characteristics.

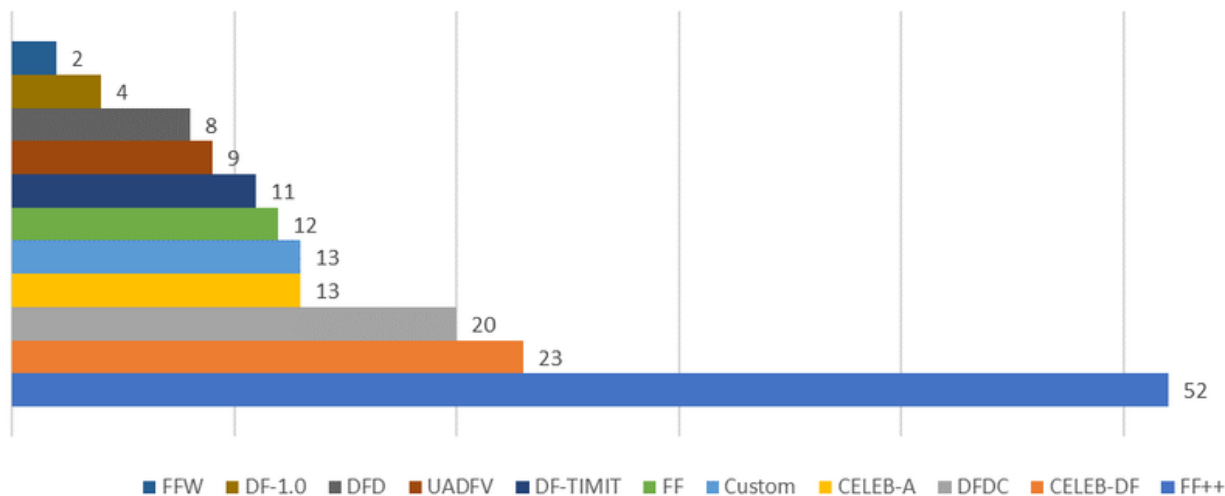

**Fig. 7.** The frequency of used datasets in recent researches.

## 4.2 Fake Datasets

- The **UADFV** is the dataset from Li et al.[48]. It has 98 videos, 49 of which are real YouTube videos, and the number of fake videos created with the help of FakeApp is 49.

- Korshunov et al.[49] made a database called **Deepfake-TIMIT** in 2018. From 32 different subjects, It has 620 videos which are Deepfake. There are 20 different Deepfake videos in each dataset. Ten videos are 128 x 128 in size, while the others are 64 x 64 in size. Faceswap-GAN is used to make the synthesized videos.

- **DFDC Preview** is a dataset in which 5000 videos generated with two facial manipulation algorithms.Dolhansky et al.[50] The character of said dataset are of different genders, skin tones, ages, etc. Video participants were allowed to record their videos with any background they desired, which resulted in visually diverse backgrounds. Videos include a variety of lighting conditions and head poses.

- **Google DFD** Dufour and Gully et al.[51] More than 3000 tampered videos featuring 28 characters in various scenes. Deepfake generation techniques are used to create the videos from 100s of real videos that are freely available on the internet.

- **FaceForensics++** is a popular dataset that consists of one thousand real video sequences and four thousand videos Rossler et al. [52] that were generated by four automatic face manipulation algorithms as Deepfakes, Face2Face, FaceSwap, and NeuralTexture A total of 977 traceable YouTube videos were used in the creation of the videos. The majority of people in the videos have frontal faces.

- **Celeb-DF** comes with a tremendous challenging video dataset of Deepfake [53], Celeb-Df used an improved synthesis method which generate around 5,639 high-quality videos.Its also have 590 original videos collected from YouTube with different ages, ethic groups and genders.

- **DeeperForensics-1.0** The largest dataset for face forgery detection has been made available Jiang et al. [54]; it consists of 60,000 videos with 17.6 million frames. There are 100 actors included in the 10,000 edited and 50,000 raw videos. There are 55 men and 45 women among them. The actors' skin tones range from black, brown, yellow, and white. Without glasses or decorations, all faces are clean in it.

- **Wild Deepfake** Zi et al. [55] The dataset Wild Deepfake , is made up of 7,314 face sequences takenfrom 707 Deepfake videos that were all found on the internet. By testing Deepfake detectors against real-life Deepfake , WildDeepfake is able to evaluate their effectiveness.

- **OpenForensics** On multi-person images, Le et al. [56] takes care of Deepfake's detection capabilities. With almost 334,126 faces and a total of 115,325 images (45,473 original photos and 70,325 manipulated images). OpenForensics can also be utilized for segmentation and tasks standard object detection, demonstrating its versatility. It also includes facial information such as facial landmarks, a forgery category, a segmentation mask, and a bounding box.

- The **Diverse Fake Face Dataset (DFFD)** Dang et al. [57] have created a large dataset of face counterfeit. Males account for 47.7 percent of all photographs or videos, 52.37 percent are female, with an averageage of 21-50 years. The real face samples are taken from FFHQ, CelebA, and FaceForensics++. Video clips from FaceForensics++ were used for facial identification and expression swapping. In order to generate attribute manipulated images, they have used two techniques:StarGAN[17] and FaceAPP[12]. In recent years, PGGAN and StyleGAN [58] have been used to synthesize face images.

- The **GBDF Gender Balanced Deepfake Dataset**[59] dataset is created using FF++(c23 version), Celeb-DF, Deeper Forensics-1.0 and consist of 10, 000 videos with 5000 each for males and females.

- **DFDC** Facebook has built a huge dataset of face videos that can be used to train detection models. A well-known dataset Dolhansky et al. [60] Deepfake Detection Challenge (DFDC) Kaggle competition was arranged by them. There are 128,514 videos in DFDC, around 3,426 actors helped to create 100,000.

- **MICC: F220 , F2000 , F600** [61]Images from the MICC-F600, MICCF2000, and MICC-F220 datasets are used to spot copy-move modifications. In MICC-F220, there are 110 altered and 110 original photographs; there are 700 tampered and 1300 authentic images in MICC-F2000. Moreover, in MICC-F600, there are 160 modified and 440 raw images.

  ○ Table 2: Details of publicly available data set for Deepfake

| Table 2 | | | | | | |
|---------|---------|---------|---------|---------|---------|---------|
| Year | Dataset | Original | | Fake | | Details of Dataset |
| | | Image | Videos | Image | Videos | |
| 2011 | ▪ MICC [61] | 1300 | | 700 | | ▪ Used for image copy-move tampering detection. |
| 2015 | ▪ WWD [62] | 13,500 | | | | 92 variations of forgery, 82 instances of forgery, and 101 different splice detection masks. |

| 2015 | ▪ CelebA [47] | 202,000 | | | | This dataset of photos includes a wide range of poses and cluttered backgrounds. |
| 2017 | ▪ VISION [63] | 34,400 | 1914 | | | A video and image source identification application-based dataset (35 portable devices of 11 major brands). |
| 2018 | ▪ UADFV [48] | 17,300 | 49 | 17,300 | 49 | The FakeAPP is used to create the Deepfake videos; it is simple to use and only contains two classes: real and fake. |

| 2018 | ▪ DF-TIMIT [49] | 34,000 | 320 | 68,000 | 640 | Low-quality(LQ) and high-quality DFTIMIT( HQ), obtained using a Faceswap GAN model. |
| 2019 | ▪ FF [64] | 500,000 | 1004 | 521,400 | | Two ways to generate Deepfake: Face2Face, and self-reenactment. |
| 2019 | ▪ FF++ [52] | 509,900 | 1,000 | 509,000 | 4000 | Face2Face and FaceSwap are two graphics-based methods, and two learning-based methods (Deepfakes and Neural Textures). |
| 2019 | ▪ DFFD [57] | 58,700 | 1,000 | 240,300 | 3,000 | Several forgery types are combined in the DFFD dataset. |

| 2019 | ▪ DFD [51] | 315,400 | 363 | 2,242.7k | 3,068 | Google contributed with the FF++ Dataset; additionally actors, to create maipulated videos |
| 2019 | ▪ DFDC-P [50] | 488,400 | 1,131 | 1,783,300 | 4,113 | Featuring two facial modification algorithm |
| 2020 | ▪ DFDC [60] | | 23k | | 104k | To expand the DFDC-P, 8 facial modification algorithms have been applied. |
| 2020 | ▪ Celeb-DF [53] | 225,400 | 590 | 2,116,800 | 5,639 | 59 celebrity YouTube videos were created utilising an upgraded Deepfake synthesis technique. |
| 2020 | ▪ DF-1.0 [54] | 12.6M | 50,000 | 5.0M | 10,000 | A large-scale dataset for real-world face forgery detection. |

| 2020 | ▪ WDF [55] | 11.8M | | 7,314 | 707 | A modest dataset that examines the real-world dataset's ability to recognise Deepfakes. |
|---|---|---|---|---|---|---|
| 2022 | ▪ GBDF [59] | | 7500 | | 2500 | It consist of 10,000 videos with 5000 each for males and females with 1:4 real to fake ratio. |
| 2021 | ▪ OF [56] | 16,000 | | 173,000 | | Huge challenging dataset for segmentation in the wild and multi-face forgery detection. |

- **WWD** The Wild Web Dataset (WWD) [62] contains 101 distinct mask splice detections and 82 instances of 92 forging variations. The WWD aims to close a gap in the evaluation of localization techniques for image alteration.
- **VISION** A VISION dataset that contains 648 raw videos and 11,732 original images. [63] The pictures were posted on social websites such as WhatsApp and Facebook, while the videos were distributed on WhatsApp and YouTube, generating 1,914 videos and 34,427 images.
- **FaceFornesics (FF)** [64] dataset is based on DeepFake aims to carry out forensic operations on manipulated images, such as segmentation and facial recognition. Over 500,000 frames from 1004 (facial videos taken from YouTube) are included.

# 5 Deepfake Detection

## 5.1 Fake Image Detection

### Spatial based Detection

Afchar et al.[65] trained CNN classifiers by deepfake online video and Face2Face dataset. Two types of Meso-4, Mesoscopic and MesoInception-4, have been presented in it. In detection of fake videos the accuracy was 98 percent and 95 percent achieved respectively. Nataraj et al.[66] purposed a co-occurrence matrices on the bases of more than 56,000 RGB images, generated through cycleGAN and StarGAN, and achieved 99.71 percent and 99.37 percent accuracy respectively. Another method based on inter-frame and temporal inconsistencies used by Güera et al.[67] through CNN and LSTM models, trained by a collection of 600 videos obtained from multiple websites. The highest accuracy of the model is greater than 97 percent. Nguyen et al. [68] used **VGG-19, Capsule networks on Deepfake online videos, FF, replay-attack** datasets and achieved accuracy 96.52 percent, 94.47 percent and 99.13 percent respectively moreover, presented an FDFtNet, a novel detector to increase the capacity of current CNN models like SqueezeNet, ShallowNetV3, ResNetV2, and Xception. FDFtNet achieves an overall accuracy of 90.29 percent in detecting fake images generated from the GANs-based dataset, outperforming the state-of-the-art. Jeon et al.[69] used PGGAN, DF, and FF datasets, and obtained 90.29 percent, 97.02 percent, 96.67 percent accuracy respectively. Using universal texture data, researchers enhanced the validity and adaptability capacities of existing CNNs in identifying generated fraudulent faces. Fung et al.[70] used FF++, UADFV and Celeb-DF datasets with Xception network, SVM, and Bayes classifier technique and found high accuracy 99.7 percent, 96.8 percent, 90.5 percent. Li and Lyu. [71] embraced a model based on deep learning capable of distinguishing between Deepfake and real videos. Furthermore, VGG-16, ResNet with UADFV, and DF-TIMIT datasets were prepared using the Face-warping artifacts feature. It gained 97.4 percent and 93.2 percent accuracy. Using CelebA, LSUN bedroom scene datasets with Image fingerprint feature, Yu et al.[72] achieved 99.43 percent, 98.58 percent accuracy.

Fine-tuning is a powerful strategy to protect the DNN model from adverse disruptions in fingerprint pictures. According to Chai et al. [73], duplicative artifacts were assessed from local patches in order to find the fake face. This theory has been tried using various available techniques, such as CNN [74], Xception [75], MesoInception4 [65], and Resnet18 [76] having p values 0.5 and 0.1 using CelebA-HQ dataset.

Matern et al. [77] adopted visual features and fake detection methods founded on elementary visual elements, like missing reflections, eye color, and missing details in the dental and eye regions. He included two distinct classifiers in this case study: 1st is the Logistic regression model, and the 2nd is Multilayer Perceptron (MLP). The own dataset was used to train the model, and 85 percent accuracy was achieved. Yang et al. [78] presented Head Pose Features for detecting the Deepfake efficiently. SVM classifier trained by Own dataset like: UADFV, Deepfake TIMIT (LQ) Deepfake TIMIT (HQ), FF++ / DFD, DFDC Preview, Celeb-DF, used for detection and result obtained upto 89 percent.

Bharati et al. [79]proposed a novel supervised deep Boltzmann machine algorithm.The proposed approach for classifying images as original or retouched yields an accuracy of over 99 percent on three makeup data sets[17]. Rossler¨ et al.[52] came up with the Mesoscopic, Steganalysis, and Deep Learning Features. The detection system was totally based on a convolutional neural network classifier that was trained on FF++ (Face2Face, RAW), FF++ (NeuralTextures, RAW) datasets.. The maximum accuracy was approximately 99 to 100 percent.

Table 3: Spatial based techniques used for deepfake detection

| Table 3 | | | | | |
|---|---|---|---|---|---|
| Year | Study | Methods | Techniques | Accuracy | Dataset |
| 2022 | [80] | Hybrid Image Transformer | Dual CNN | 98.57% | FaceForensics++ |
| 2018 | [65] | Meso-4, MesoInception-4 | CNN | 95% & 98% | Deepfake online videos, FF. |
| 2019 | [66] | Co-occurrence matrices | DNNs | 99.71% &99.37% | Own(CycleGAN and StarGAN) datasets |
| 2018 | [67] | inter-frame and temporal inconsistencies | CNN, LSTM | 97% | A bundle of 600 videos gathered from various websites. |
| 2019 | [68] | Capsule-forensics | VGG-19, Capsule networks | 94.47% & 99.13% | Deepfake online videos, FF, REPLAYATTACK database |
| 2020 | [69] | Fine-Tune and transformer | SqueezeNet, ShallowNet,ResNet, Xception | 90.29%, 97.02%, 96.67% | PGGAN, DF, FF |
| 2021 | [70] | Unsupervised Contrastive Learning | Xception network, SVM, and Bayes classifier | 99.7%, 96.8%, 90.5% | FF++, UADFV and Celeb-DF. |
| 2018 | [71] | Face-warping artifacts | VGG-16, ResNet | 97.4%, 93.2% | UADFV and DF-TIMIT |

| 2019 | [72] | Image fingerprint | DNN | 99.43%, 98.58% | CelebA, LSUN bedroom scene dataset |
|------|------|------------------|-----|----------------|----------------------------------|
| 2020 | [73] | Patch-based classification | Resnet-18, Xception, MesoInception4 | 100% | CelebA-HQ |
| 2019 | [77] | Visual Features | Logistic Regression, MLP | 85% | Customized dataset (DFDC Preview, DeepfakeTIMIT LQ/HQ, Celeb-DF , UADFV, FF++/DFD ) |
| 2019 | [78] | Head Pose Features | SVM | 89% | Customized dataset (DFDC Preview, DeepfakeTIMIT LQ/HQ, Celeb-DF , UADFV, FF++/DFD ) |
| 2019 | [52] | Mesoscopic ,Steganalysis and Deep Learning Features | CNN | 100%, 99% | FF++ (Face2Face, RAW), FF++ (NeuralTextures, RAW) |
| 2020 | [56] | Deep Learning Features | CNN + Attention Mechanism | 99% | FF++ (Face2Face, -) |

Khan et al. [80] proposed hybrid image transformation using XceptionNet and EfficientNet-B4. The Faceforeincis ++ dataset was used to train the model and achieved 98.57 percent accuracy.

## Frequency based Detection

In 2019, a generic fake face photodetection technique was presented by Xuan et al. [81]. DCGAN, WGAN-GP, PGGAN trained by CelebA-HQ dataset. The primary goal was to explicitly include a preprocessing stage in the training process for removing petty instability artifacts in GAN images, which drives the forensics classifier to concentrate on higher intrinsic forensic indications for recognizing GAN-based pictures. The highest accuracy was 95.45 percent. Another method based on Temporal discrepancies was created by Sabir et al.[82]. CNN and RNN models were trained by FF++ dataset to detect the Deepfake. In this analysis, only low-quality videos were included and 96.9 percent accuracy was gained. Jeon et al. [83] came up with a self-training method and trained EfficientNet and ResNext architectures by their own (TPGGAN and StyleGAN) based dataset, which efficiently detects Deepfake images with the accuracy of 98.49 percent.

Best results have been recently addressed by Guarnera et al.[84] presented a fake detection model. This model is based on the research of convolutional imprints. Expectation Maximization Algorithm [85] was used to extract features. K-NN, SVM, and LDA classifiers were trained by Own(AttGAN, GDWCT,StarGAN, StyleGAN, StyleGAN2) dataset for best detection. Final result was 99.81 percent accurate.

Nirkin et al. [86] identified the distinctions between real faces and their setting to identify unreal faces. The first network is prepared to determine an individual's face with FF++, Celeb-DF, DFDC datasets, whereas the second context recognition network considers the context of the person's face. 99.7 percent, 66.0 percent, 75 percent accuracy was achieved using.

McCloskey et al. [87] presented a detection approach subject to color components and a Linear Support Vector Machine (SVM). It was proposed for the GAN-Pipeline Features technique, having a detection method based on color characteristics and a linear SVM for the conclusive categorization with NIST MFC2018 dataset. The writers assumed that the color differs between fake synthesis pictures and real camera photographs. While evaluating 70.0 percent accuracy was achieved. Marra et al. [88] adopted unique Deep Learning Features for the best detection. Authors suggested a multi-task incremental learning detection method for detecting and classifying new types of GAN generated images while maintaining past performance. Own (CycleGAN, ProGAN,Glow, StarGAN, StyleGAN) dataset was used for model training and achieved 99.3 percent accuracy.

Dang et al. [57] comprehensively analyzed different facial manipulations. The authors suggested using attention mechanisms and famous CNN models such as Xception- Net and VGG16 using a dataset based on DFFD (ProGAN, StyleGAN).Final accuracy was 100 percent. Hulzebosch et al. [89] adopted Deep Learning Features and considered different scenarios like cross-data, cross-model, and Post-processing. Fake detection techniques were based entirely on the renowned Xception network along with ForensicTransfer [90], which is basically an Autoencoder method. CNN, AE classifiers were trained by their own (StarGAN, Glow, ProGAN, StyleGAN) based dataset to detect fake images and got 99.8 percent accuracy, which is best performance.

A deep learning technique based on Restricted Boltzmann Machine (RBM) was proposed by Bharati et al. [79] in order to identify digital editing of face pictures. As a concerned database, the writers designed two fake databases from the actual ND-IIITD database [91] and a group of celebrity face photos that were downloaded from the internet. Their approach achieved 96.20 percent accuracy respectively.

Zhang et al.[92] presented a spectrum domain component to witness faux images. It was trained on its own (StarGAN/CycleGAN) dataset. Regarding the classifier, AutoGAN was put forward; it's a GAN simulator capable of making GAN artifacts in any photo without using an already trained GAN model. Ultimately, the accuracy obtained was 100 percent.

Table 4: Frequency based techniques used for deepfake detection

| Table 4 | | | | | |
| Year | Study | Methods | Techniques | Accuracy | Dataset |
|---|---|---|---|---|---|
| 2019 | [81] | Preprocessing combined with deep network | DCGAN, WGAN-GP, PG-GAN | 95.45% | CelebA-HQ |
| 2019 | [82] | Temporal discrepancies | CNN and RNN | 96.9, 94.35, 96.3% | FF++ |
| 2020 | [83] | Self-training | EfficientNet and ResNext | 98.49% | Customized dataset (StyleGAN and TPGGAN) |
| 2021 | [86] | Discrepancies between two regions | Xception networks | 99.7%, 66.0%, 75% | FF++, Celeb-DF, DFDC |
| 2018 | [87] | GAN-Pipeline | SVM | 70.00% | NIST MFC2018 |
| 2020 | [89] | GAN-Pipeline | k-NN, SVM, LDA | 99.81% | Own(AttGAN, GDWCT,StarGAN, StyleGAN, StyleGAN2) |
| 2019 | [88] | Deep Learning Features | CNN + Incremental Learning | 99.30% | Own(CycleGAN, ProGAN,Glow, StarGAN, StyleGAN) |
| 2020 | [57] | Deep Learning Features | CNN + Attention Mechanism | 100.00% | DFFD (ProGAN, StyleGAN) |
| 2020 | [89] | Deep Learning | CNN, AE | 99.80% | Own(StarGAN,Pro GAN, StyleGAN) |
| 2016 | [79] | Deep Learning Features (Face Patches) | RBM | 96.20% | Own(Celebrity Retouching, ND-IIITD Retouching) |

| 2019 | [92] | Spectrum Domain Features | GAN Discriminator | 100.00% | Own StarGAN/CycleGAN based |
|---|---|---|---|---|---|

## 5.2 Fake Video Detection

Lip Forensics is a method for detecting face forgeries in videos that is generalizable and robust. Haliassos et al. [93] used Semantic irregularities methods and used FaceForensics++, DF-1.0, Celeb-DF, DFDC datasets with the ResNet-18 technique and obtained the accuracy of 99.7 percent, 82.4 percent, 73.5 percent respectively. Tariq et al.[94] used FF++, DFD and Deepfake Wild videos datasets on Convolutional LSTM-based Residual Network (CLRNet) and got 97.50 percent, 97.13 percent, 97.23 percent accuracy. The fundamental concept is to use a convolutional LSTM-based residual network (CLRNet) with a unique training technique to track the spatial and temporal information in Deepfakes.

In [95] Deepfake detection, a blend of stationary biometrics on facial recognition and secular behavioral biometrics on facial sentiments and head motions was presented. In addition, their focus was on features, appearance and behavior so they used ResNet-101, VGG technique utilizing DFD, Celeb-DF, DFDC, and FF++ datasets and achieved 97.7 percent, 98.9 percent, 93.2 percent, 96.45 percent accuracy. Agarwal and Farid [96] proposed a detection system based on both facial expressions and head movements, using an SVM (Support Vector Machine) classifier trained over the Own (FaceSwap, HQ) dataset to detect the fake videos. The accuracy was achieved 96.3

Mittal et al.[97] utilized the Siamese network to simulate the visuals and audio in movies as well as trained the Siamese network with DF-TIMIT and DFDC datasets and gained 94.9 percent, 84.4 percent accuracy. Furthermore, Chugh et al. [98]also used DFDC and DF-TIMIT datasets with MDS(Modality Dissonance Score) network technique, to obtain 91.54 percent and 94.7 percent accuracy. Evolutionary divergence results are computed between audiovisual segments across one-second video gaps, and the MDS is ascertained after aggregating all parts.

Agarwal et al.[42] Proposing an approach for detecting fraudulent videos based on irregularities in the dynamics of the mouth shape (visemes) and the prominent phonetic even more **Phoneme-viseme mismatches** technique was used through CNN along with A2V, T2V datasets OpenFace2 toolkit OpenFace 2.0: Facial Behavior Analysis Toolkit [99] was considered for feature extraction. and gained 96.9 percent, 71.1 percent, accuracy. Qi et al.[100] adopted the Heartbeat rhythms method to highlight cardiac rhythm signals; the scientists built a motion-magnified spatial-temporal representation (MMSTR) for the video. After using DeepRhythm technique with FF++, DFDC-P datasets, they obtain 99.7 percent, 74.5 percent results.

Table 5: List of techniques and methods used for fake video detection

| Table 5 |
|---|

| Year | Study | Methods | Techniques | Accuracy | Dataset |
|------|-------|---------|-----------|----------|---------|
| 2021 | [93] | Semantic irregularities | ResNet-18 | 99.7%, 82.4%, 73.5% | FF++, DFDC, Celeb-DF, DF-1.0 |
| 2021 | [94] | Spatial and temporal information | CLRNet | 97.50%, 97.13%, 97.23% | DFD, FF++, Deepfake -in-the-Wild videos |
| 2018 | [101] | Eye blinking | CNN, LRCN | 98%, 99% | For LRCN EBV dataset and For CNN CEW Dataset |
| 2020 | [95] | Using appearance and behavior | ResNet-101, VGG | 97.7%, 98.9%, 93.2% | The world leaders dataset along DFDC, DFD, FF++ and Celeb-DF. |
| 2020 | [97] | Using emotion audio-visual affective | Siamese network architecture | 94.9%, 84.4% | DF-TIMIT and DFDC. |
| 2020 | [102] | Phoneme-viseme mismatches | CNN | 96.9%, 71.1% 80.7% | A2V, T2V |
| 2020 | [98] | Modality Dissonance Score | MDS network | 91.54%, 94.7% | DFDC, DF-TIMIT |
| 2020 | [100] | Heartbeat rhythms | DeepRhythm | 99.7%, 74.5% | FF++, DFDC-P |
| 2021 | [103] | Consistency of eyes and gazes | 3 dense layers network architecture | 93.28%, 80.00% | FF++, CelebDF |
| 2019 | [96] | Head Pose and Facial Features | SVM | 96.30% | Own (FaceSwap, HQ) |
| 2020 | [104] | Eye Blinking | Distance | 87.50% | Own |

| 2018 | [105] | Deep Learning Features+ Steganalysis Features | CNN, SVM | 85.10% | Customized dataset (DFDC Preview, DeepfakeTIMIT LQ/HQ, Celeb-DF, UADFV, FF++/DFD ) |
| 2019 | [106] | Deep Learning Features | AE + Multi-Task Learning | 76.30% | FF++ and DFD |
| 2019 | [79] | Deep Learning Features | CNN + AM | 99.40% | DFFD |
| 2020 | [57] | Facial Regions Features | CNN | 100%, 99.4%, 91%, 83.6% | DFDC Preview, UADFV, FF++ (HQ, FaceSwap), Celeb-DF |

In [101] three steps were involved in the composition of the LRCM model: (i) Characteristic extraction from the eye series using VGG16, (ii) Sequence comprehension through LSTM, and (iii) State prediction, which yields the probability of eye open and closed conditions on the basis of the result of LSTM. Li et al. [101] used CEW Dataset and Eye blinking method for EBV and CNN dataset for the LRCN system and attained 98 percent and 99 percent accuracy. Furthermore, Jung et al. [104] developed a system known as DeepVision to examine fluctuations in blinking patterns. Fast-HyperFace and Eye-Aspect-Ratio (EAR) were used together to find the face and figure out the eye aspect ratio. Finally features based on the number of blinks and the time of the blinks were encountered to determine whether the video was fake or real. This technique had an absolute accuracy rate of 87.5 percent over its own dataset.

Zhou et al. [105] proposed Deep Learning Features and Steganalysis Features methods with convolutional neural network and SVM classifier using UADFV dataset and a final accuracy rate of 85.1 percent was obtained.

Nguyen et al. [106] presented a Capsule Networks-based fake detection model. This system had relatively fewer parameters than a standard CNN; however, it had the same outcomes. A detection system that used an auto-encoder was considered. Final accuracy was 76.3 percent. Nguyen et al. [68] proposed a new fake detection system based on Capsule Networks. This method had fewer parameters than a traditional CNN with the same results. The proposed detection system was tested with the FaceForensics++ dataset, which had accuracy rates of more than 96 percent. Dang et al. [57] came up with the convolutional neural network and Attention Mechanism classifier using the DFFD dataset. In the case of detecting identity swaps, the proposed method had an accuracy of 99.43 percent.

Guerra et al . [67] proposed video temporal features that were capable of telling if a frame is manipulated or not with a combination of CNNs and RNNs model. The authors utilized InceptionV3 which had been pre-trained with the ImageNet dataset for the CNN and LSTM model with one hidden layer and 2048 memory blocks that could be used for the RNN system. The accuracy was 97.1 percent. Tolosana et al. [107] came up with the Facial Regions Features method. The author considered a fake detection system based on a CNN classifier trained on DFDC Preview, Celeb-DF datasets. The result obtained was 91.0 percent, 83.6 percent respectively.

## 5.3 Methodologies for Deepfake Detection

In this part, we will look at various approaches to Deepfake detection. The PRNU (Photo Response Non Uniformity) pattern of a digital image, is a noise pattern created by small factory defects in the light-sensitive sensors of a digital camera. This noise pattern is highly individualized and often referred to as the fingerprint of the digital image. PRNU analysis considers a method of interest because it expects that manipulating the facial area will affect the local PRNU pattern in the video frames [108].

Face tempting videos are also automatically detected using Meso-4 and Misconception-4 [19], both the networks used less number of layers for data. As a result of this study, an accuracy of 98 percent and 95 percent have been achieved.

Nataraj et al. [66] used a steg analysis method to find fake images. The co-occurrence matrices methods were made from RGB images, and the values were trained with a deep convolutional neural network to differentiate the real ones from the fake ones. RGB images were used to make the co-occurrence matrices methods. Moreover, the values were prepared with a deep convolutional neural network (CNN) to distinguish the real from the fake ones. In [57] StarGAN and CycleGAN-based forged photos could be classified, achieving 99 percent valid results. Li et al. [109] have looked at the statistical properties of deep network-generated pictures, like the connection between bordering pixels in HSV and YCbCr color spaces, in order to see if the difference between real and fake images can be detected using DNNs with the linear discriminative technique using dataset LFW, LSUN, FFHQ, CelebA, FFHQ.

There is another generalizable and robust technique lip forensics which is used for detecting face forgeries Haliassos et al. [93] propose to utilize the high-level semantic irregularities using face forensics ++, DF-1.0, Celeb-DF, DFDC datasets with the ResNet-18 model and obtained this accuracy 99.7 percent, 82.4 percent, 73.5 percent.

Lugstein et al. [110] came up with a new way to tell whether data is a Deepfake or not, based on how different the photo response no uniformity (PRNU) is. So, the PRNU and SVM technique is well-known because facial retouching and face morphing attacks can be detected.

Guerra et al. [67]provided CNN and LSTM detection method which is based on face-swapping. In this method, InceptionV3 which is CNN is used to gather frame-level features, those features passed to LSTM to build a sequence descriptor for classification and 97.1 percent accuracy was achieved.

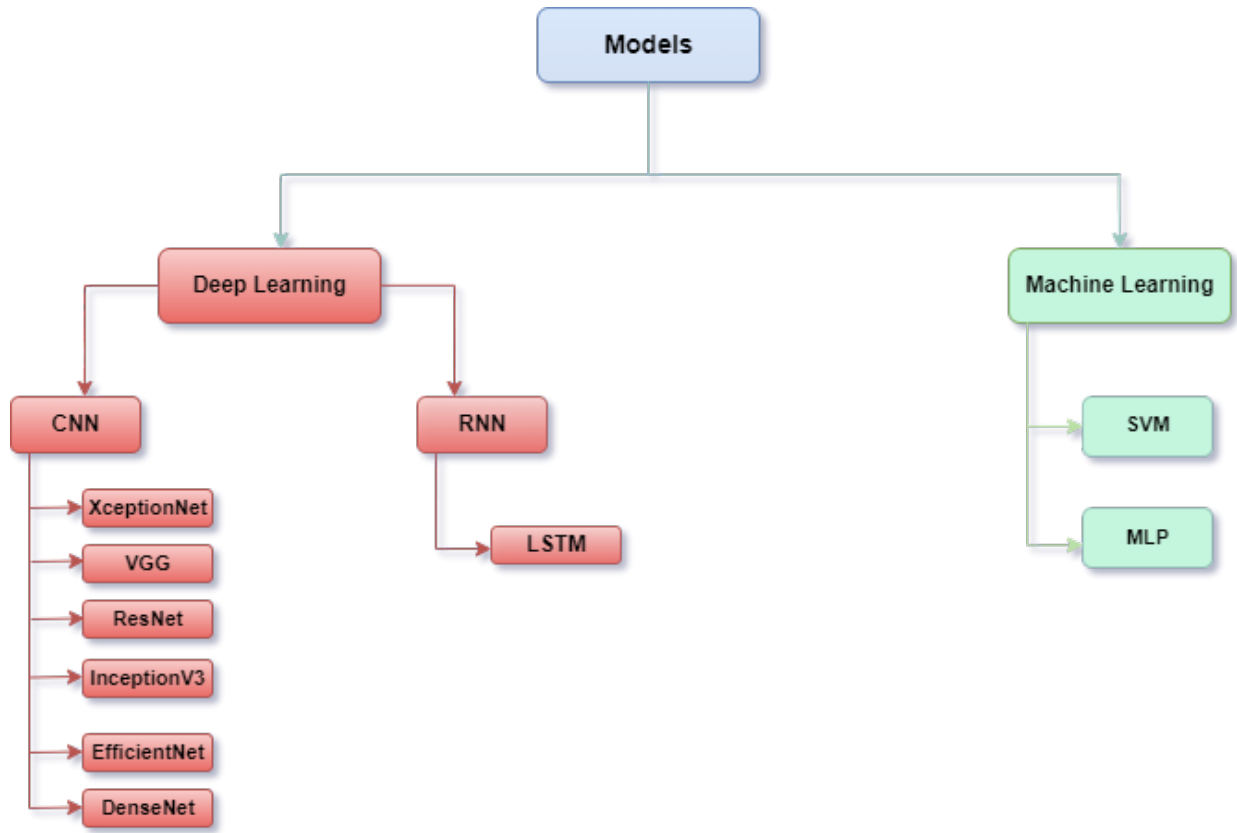

**Fig. 8.** List of models used for deepfake detection in current study

Nguyen et al.[68] operated VGG-19 along with Capsule networks in order to recognize fake images and videos, including computer-generated image detection and replay attack detection using these techniques with dataset Deepfake online videos and FF and got the accuracy of 94.47 percent, 99.13 percent.

Xuan et al.[81] used Preprocessing combined with a deep network method with the DCGAN, WGAN-GP, and PGGAN techniques. The intention was to dismiss low-level, inconsistent artifacts of GAN visuals to make the forensics classifier prioritize higher innate forensic indicators so that GAN-based images could be detected The 95.45 percent accuracy was achieved . Sabir et al. [82] created a method based on Temporal discrepancies, using CNN and RNN technique on the video to recognize, trim, and align faces in a video. The model used micro-, meso-, and macroscopic features for detection. These landmark-based components produce facial alignment with recurring directional DenseNet that excels at spotting face-manipulated videos. With CNN and RNN models trained on FF++ dataset,accuracy was gained 96.9 percent, 94.35 percent, 96.3 percent. Jeon et al. [69] created CNN models like Squeeze Net, ShallowNetV3, ResNet2, and Xception more powerful. The fine-tuning technique is employed to get the characteristics out of MBblockV3 whereas PGGAN, DF, FF datasets are used for training . This approach is a fine-tuning transformation. It obtained 90.29 percent, 97.02 percent, 96.67 percent accuracy. Jeon et al. [83] ddevised a GAN-image detection framework (T-GD) to assemble a model trained on TPGGAN and StyleGAN- based datasets that quickly and accurately detects Deepfake images using Efficient Net and ResNext techniques. The model is based on the relationship between the teacher and the students, improving the detection performance. It obtained 87.80 percent and 98.49 percent accuracy.

Hsu et al. [111] developed a couple-wise learning model capable of telling if a GAN-generated fake image is real. He merged an enhanced model of the DenseNet backbone network with an entirely different network called the Siamese network to make a unique model. That model is known as Common Fake Feature Network (CFFN).

Jain et al.[112] used an adversarial perturbation method to improve the performance of existing Deepfake models.DIP - deep image prior - and Lipschitz regularization are used to improve the validity of CNN-based deepfake sensors. DIP defence achieved 98 percent accuracy. Wu et al. [113] developed a system called SSTNet that integrates spatial, steganography, and characteristic extraction techniques to determine Deepfakes. In this method, XceptionNet monitors the image's spatial features and statistics. Also, to mine temporal characteristics, RNN is used. This extracted data is applied for binary classification for identifying Deepfakes. Global texture data can be analyzed to improve the existing CNNs by making them more robust and generalized when they are used to identify fake faces. The gram-Net technique evaluates JPEG compression, downsampling, noise, and blur. Gram-Net has proven to work with a wide range of GANs. Conclusively, using the Analyzing global image texture method, 95.51 percent and 90 percent accuracy was achieved with the help of the ResNet model technique by usingCelebA-HQ and FFHQ images datasets[114]

To differentiate between real and fake images, Khalid et al.[115] created the OC-FakeDect technique that utilizes a one-class Variation Auto Encoder (VAE) and solely trains on real face photos.

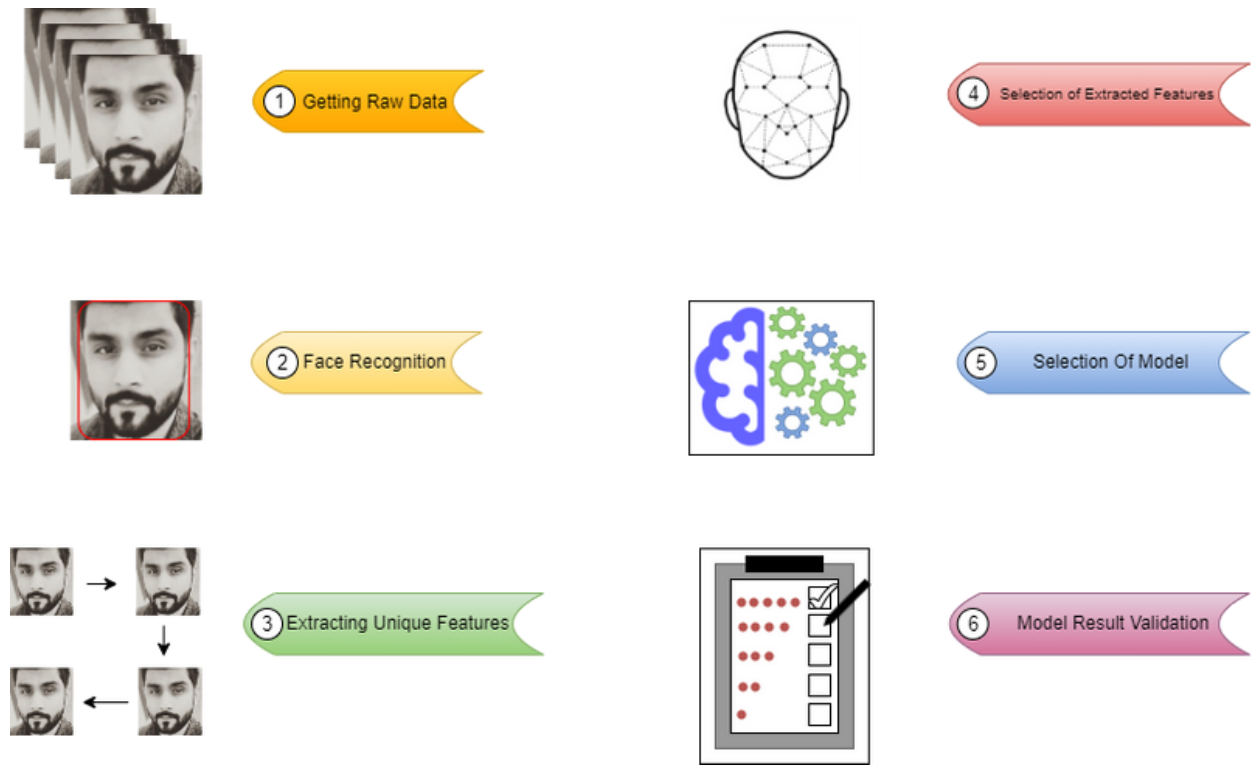

**Fig. 9.** Generic steps used in deepfake detection

.

Fung et al. [70] came up with an unsupervised way to tell when someone's face is changed. It is a new way to detect unlabeled data to see if someone has changed their face. When a face image creates an image, it goes through two different transformations and into two different subnetworks, Xception and projection head network. When they used two different transformations, an FF++ and UADFV, as well as Celeb-DF and Celeb-DF datasets, were performed with the Xception network, SVM, and Bayes techniques; it appeared that they all had high accuracy.

Tariq et al. [94] came up with a way to find many different types of Deepfakes. To analyze the spatial and secular data in Deepfakes, they use a convolutional LSTM-based residual network (CLRNet) approach qualified on trained DFD, FF++, and Deepfake-in-the-Wild video datasets. Also, the model tested on Deepfakes that had never been seen before, it got 97.50 percent, 97.13 percent, and 97.23 percent accuracy.

In less than a minute, this app can alter faces. For this purpose, the essential components of the face image are determined and indicated as descriptors. Because every key point is distinctive, one must use a second clustering procedure to make the codebook per picture.[116].

Using a CNN and an RNN, Li et al. [101] devised a method based on eye blinking for detecting fake faces in videos. With VGG16, a type of RNN, features were extracted from the eye sequence using an LRCN model.

Sequence learning will be done using LSTM, predicting whether the eye is open or closed. CEW and LRCN datasets achieved 98 percent and 99 percent accuracy. Similarly, a unique deep learning-based model was introduced that was capable of distinguishing real videos from fake ones. The model employs the warping action that is used to create Deepfake. There is a gap in the resolution between the warped face area and the rest of the image, creating artefacts. First, CNN is prepared to identify faces. Then, it attempts to find landmarks and calculates transformed matrices to ensure the faces are in the right place. UADFV and DF-TIMIT datasets were used with VGG-16, ResNet technique and 97.4 percent, 93.2 percent accuracy was faced.

Agarwal et al.[96] proposed a detection system based on Head Pose and Facial Features method. For the classification, SVM was used.

McCloskey et al. [87] designed a GAN generator's architecture, making it easier to find visual flaws in Deepfake images. in Deepfake visuals. The generator's normalizing techniques are adopted, which reduce the total pixels that are overexposed or underexposed. A detection system subject to color attributes and a linear Support Vector Machine (SVM) for the ultimate classification gained 70.0 percent accuracy. Yu et al. [72] looked at GAN fingerprints for image attribution and used them to tell whether images were real or made by GANs. If the model is trained with slight modifications in the dataset, its fingerprint would be distinct, making it easier to identify the model. Fine-tuning is also a good way to protect the DNN model from changes in fingerprint images that aren't good for it. CelebA, LSUN bedroom scene datasets with Image fingerprint feature achieved 99.43 percent, 98.58 percent accuracy. Yu et al. [78] created GAN fingerprints to attribute images with SVM classification using Deepfake TIMIT (LQ) Deepfake TIMIT (HQ), UADFV, Celeb-DF , FF++ / DFD, DFDC Preview datasets found 89.0 percent, 55.1 percent, 53.2 percent, 47.3 percent, 55.9 percent, 54.6 percent results.

Matern et al. [77] utilized Deepfake and face manipulation tools based on visual characteristics such as teeth, eyes, and facial features. The method was proposed by adopting the Visual Features method and Fake detection methods based on very simple visual characteristics such as eye colour, missing reflections, and missing details in the eye and dental areas. Deepfake TIMIT (LQ) Deepfake TIMIT (HQ), UADFV, Celeb-DF , FF++ / DFD, DFDC Preview datasets used for the training and 70.2 percent, 77.0 percent, 77.3 percent, 78.0 percent, 66.2 percent, 55.1 percent accuracy achieved in the testing.

Fernandez et al. [117] used a heartbeat rhythm and designed a Deep Rhythm model that could tell if a video was fake. For the purpose of showing heart rhythm signals, A motion-magnified spatial-temporal representation (MMSTR) model was used to make the video. As a result, a dual-spatial-temporal attentional network was produced to catch if a video was fake.

There are many ways to find digital fingerprints in images that can be learned and used to tell if they were made by a GAN or not[95]. CNN's are used to examine pixel-level artefacts when a face region moves onto a target. A convolutional neural network with only a few layers is called MesoNet. It can use to build a model of a microscopic object. L. Chai et al. [73] used redundant artefacts and a Patch-based classification method that can evaluate local patches. This hypothesis was assessed on the FFHQ and CelebA-HQ datasets, using recognized models like MesoInception4, Xception, Resnet-18, and CNN, having p values 0.5 and 0.1, respectively. The datasets of CelebA-HQ got 99.97 percent accuracy.

A Siamese network imitates the visuals and sounds in movies, combined with a blend of the two triplet loss functions to confine alikeness. Mittal et al.[97] trained Siamese network architecture technique with the DF-TIMIT and DFDC datasets and gained 94.9 percent and 84.4 percent accuracy .

Agarwal et al. [102] used the Phoneme-viseme mismatches technique trained on A2V, T2V datasets and proposed an approach for detecting fraudulent videos based on irregularities in the dynamics of the mouth shape-viseme mismatches) and the prominent phonetic. Some phonemes need the lips to be completely closed to be spoken correctly: A few of them are papa, mama, and baba. It gained 96.9 percent, 71.1 percent, 80.7 percent accuracy.

Deepfake videos can be detected using a method proposed Chugh et al. [98] . After the calculation of disparity scores across audiovisual segments over one second video chunk, the MDS was determined by using aggregation to all parts and taking the average.

Guarneraet. al.[84] represented a way to tell if an image is fake by looking for forensic traces that are hiddenin pictures. The EM algorithm - expectation maximization - is utilized to discover a bunch of local characteristics for modeling the underlying convolutional productive procedure. StarGAN, GDWCT, StyleGAN, StyleGAN2, and ATTGAN made Deepfakes. CelebA along with LFW datasets obtained 90.22 percent accuracy.

To find fake videos the ABC metric shown in [118] . To take into account the baseline input, the attribution methods employ Shapley values. ResNet50 trained on VGGFace2 represents the original class label, while P represents the class label that is most likely to occur.

Qi et al. [100] used the Heartbeat rhythms method to highlight cardiac rhythm signals; the scientists created an MMSTR for the video—a motion-magnified spatial-temporal representation.. After training FF++, and DFDC-P datasets using the Deep Rhythm technique, we obtained 99.7 percent and 74.5 percent accuracy. Hu et al. [119] examined the distinction between the pair of eyes while detecting Deepfake facial photos and presented a detection standard that exploits the physical boundaries in GAN-based pictures. Then it estimates the difference between both eyes to reveal whether a picture is real or fake. Demir et al. [103] ddevised a system to identify Deepfakes by examining the gaze in videos and looking for the consistency of eyes and gazes' method to analyze OR detect the video. Using three dense layers of network architecture technique with face-forensics++, CelebDF datasets, 93.28 percent, 80.00 percent accuracy were obtained.

The contrast between faces and their surroundings was exploited by Nirkin et al. [86] to identify false faces. He trained two networks: one to identify a person's face and the other to account for certain things, like hair, ears, and neck. The researchers found that discrepancies detected by comparing these two networks.

# 6 Conclusion

The study on Deepfake production and Deepfake detection has been comprehensively reviewed and analyzed in this survey, which hopefully will be useful to anyone interested in the subject. A taxonomy of diverse Deepfake evolution methods and the categorization of various Deepfake detection practices, has been offered among the most significant aspects of the technological growth of the approaches. It explains the fundamentals, advantages, and risks related to Deepfake , GAN-based Deepfake applications. Additionally, models for detecting Deepfake are addressed. In the investigations, the FF++ dataset accounts for the majority of the data points. Deep learning models (mostly CNNs) provide a large proportion of the total number of models. Deep learning-based techniques are commonly applied to detect Deepfake, as the methods are becoming increasingly popular. Using Deepfake as an experiment, the conclusion indicates that learning methods are useful in identifying Deepfake. Deepfake detection still confronts several hurdles, despite the tremendous advancements in essential multimedia technology and the development of tools and applications in recent years. This survey paper is intended to help scholars and practitioners in the respective field identify the most important research topics., and in luring more scholars to participate in this emerging and constantly expanding subject.